%% file: main.tex
\pdfoutput=1
\documentclass[11pt]{article}

\usepackage{EMNLP2023}

\usepackage{times}
\usepackage{latexsym}
\usepackage[normalem]{ulem}
\usepackage{graphicx,subcaption}
\useunder{\uline}{\ul}{}
\usepackage[T1]{fontenc}
\usepackage[utf8]{inputenc}

\usepackage{microtype}

\usepackage{inconsolata}

\usepackage{csquotes} % for quoting, can be removed
\usepackage{amsopn} % for operatorname

\title{VECHR: A Dataset for Explainable and Robust Classification of Vulnerability Type in the European Court of Human Rights}

\author{Shanshan Xu$^{1}$, Leon Staufer$^{1,2}$, Santosh T.Y.S.S$^{1}$,\\
\textbf{Oana Ichim$^{3}$, Corina Heri$^{4}$, Matthias Grabmair$^1$}\\
$^{1}$Technical University of Munich, Germany, $^{2}$LMU Munich, Germany\\
$^{3}$Graduate Institute of International and Development Studies, Switzerland\\ 
$^{4}$Faculty of Law, University of Zürich, Switzerland\\
 \texttt{\{shanshan.xu, leon.staufer, santosh.tokala, matthias.grabmair\}@tum.de} \\
  \texttt{oana.ichim@graduateinstitute.ch, corina.heri@rwi.uzh.ch}}

\begin{document}
\maketitle
\begin{abstract}
Recognizing vulnerability is crucial for understanding %unique needs 
and implementing targeted support %systems 
to empower individuals in need. 
% At the European Court of Human Rights (ECtHR), 
This is especially important at the European Court of Human Rights (ECtHR), where the court adapts convention standards to meet actual individual needs and thus to ensure effective human rights protection. 
% an institution dedicated to upholding and protecting human rights.
However, the concept of vulnerability remains elusive at the ECtHR and no prior NLP research has dealt with it.
%, and little large-scale empirical work on classifying vulnerability has been done in general. 
%To enable future research in this area, we present a novel expert-annotated multi-label dataset on the usage of vulnerability in ECtHR. We evaluate the performance of state-of-the-art models and provide comprehensive benchmark results. Furthermore, as trustworthiness is essential within the legal domain, we assess the explainability of different models on our annotated rationale dataset and observe limited agreement between model-derived explanations and experts. Given the need to address vulnerability in line with societal advancements, we also analyze the ability of models to generalize across distribution shifts. Our dataset poses unique NLP challenges and our results show that current models have significant room for improvement regarding performance, explainability and robustness. 
To enable future work in this area, we present VECHR, a novel expert-annotated multi-label dataset comprised of vulnerability type classification and explanation rationale. We benchmark the performance of state-of-the-art models on VECHR from both the prediction and explainability perspective. Our results demonstrate the challenging nature of the task with lower prediction performance and limited agreement between models and experts. We analyze the robustness of these models in dealing with out-of-domain (OOD) data and observe limited overall performance. Our dataset poses unique challenges %and our results show that current models
offering a significant room for improvement regarding performance, explainability, and robustness. %and provide comprehensive benchmark results. Furthermore, as trustworthiness is essential within the legal domain, we assess the explainability of different models on our annotated rationale dataset and observe limited agreement between model-derived explanations and experts. Given the need to address vulnerability in line with societal advancements, we also analyze the ability of models to generalize across distribution shifts. Our dataset poses unique NLP challenges and our results show that current models have significant room for improvement regarding performance, explainability and robustness. 

\end{abstract}

% \begin{figure}[h!]
%     \centering
%     \includegraphics[width=0.5\textwidth]{figure/annot-diff.png}
%     \caption{++Dummy Fig1 +++ Difference in frequency of vulnerability annotations between article 3 and non-Article 3.}
%     \label{fig:annot-diff}
% \end{figure}

%\addtolength{\parskip}{-2mm} % hmm fix this

\section{Introduction}

\begin{table*}[htpb] %htpb
  \centering
  \input{table/descrition.tex}
  \footnotesize\caption{Description of each vulnerability type. For more details, see \autoref{sec:appendix-descriptions}.}
\end{table*}
 \label{tab:description}

\input{text/introduction}
\section{Vulnerability Typology in ECtHR}

\input{text/vulnerability}

% \section{Related Work}
% \input{text/related}

\section{Data Collection and Annotations}

\input{text/data_collection}

\section{Dataset Analysis}

\input{text/dataset_analysis}

\section{Experiments}

% \begin{table}[]
% \input{table/results_cls_exp.tex}
% \caption{classification and explanation results}
% \end{table}

\subsection{Vulnerability Type Classification}
\input{text/classification}

\subsection{Vulnerability Type Explanation}
\input{text/explanation}

\subsection{Robustness to Distributional Shifts}
\input{text/cross_article}

\section{Conclusion}
We present VECHR, an ECtHR dataset consisting of 1,070 cases for vulnerability type classification and 40 cases for token-level explanation. We also release a set of baseline results,
%as a benchmark
revealing the challenges of achieving accuracy, explainability, and robustness in vulnerability classification. We hope that VECHR and the associated tasks will provide a challenging and useful resource for Legal NLP researchers to advance research on the analysis of vulnerability within ECtHR jurisprudence, ultimately contributing to effective human rights protection.

\section*{Limitations}
\input{text/limitations}

\section*{Ethics Statement}
\input{text/ethics}

% \section*{Acknowledgements}
% This document has been adapted by Yue Zhang, Ryan Cotterell and Lea Frermann from the style files used for earlier ACL and NAACL proceedings, including those for 
% ACL 2020 by Steven Bethard, Ryan Cotterell and Rui Yan,
% ACL 2019 by Douwe Kiela and Ivan Vuli\'{c},
% NAACL 2019 by Stephanie Lukin and Alla Roskovskaya, 
% ACL 2018 by Shay Cohen, Kevin Gimpel, and Wei Lu, 
% NAACL 2018 by Margaret Mitchell and Stephanie Lukin,
% Bib\TeX{} suggestions for (NA)ACL 2017/2018 from Jason Eisner,
% ACL 2017 by Dan Gildea and Min-Yen Kan, NAACL 2017 by Margaret Mitchell, 
% ACL 2012 by Maggie Li and Michael White, 
% ACL 2010 by Jing-Shin Chang and Philipp Koehn, 
% ACL 2008 by Johanna D. Moore, Simone Teufel, James Allan, and Sadaoki Furui, 
% ACL 2005 by Hwee Tou Ng and Kemal Oflazer, 
% ACL 2002 by Eugene Charniak and Dekang Lin, 
% and earlier ACL and EACL formats written by several people, including
% John Chen, Henry S. Thompson and Donald Walker.
% Additional elements were taken from the formatting instructions of the \emph{International Joint Conference on Artificial Intelligence} and the \emph{Conference on Computer Vision and Pattern Recognition}.

% Entries for the entire Anthology, followed by custom entries
\bibliographystyle{acl_natbib}
\bibliography{anthology,custom,leon}

\clearpage
\appendix

\input{text/appendix}

% \section{Label and Token Distributions}
% \label{sec:appendix-dataset}
% \input{text/appendix_dataset}

\end{document}

%% file: table/descrition.tex
\resizebox{0.98\textwidth}{!}{
\begin{tabular}{ll}
\hline
\textbf{Vulnerable Type} & \textbf{Description} \\ \hline
Dependency & \begin{tabular}[c]{@{}l@{}}Including that of minors, the elderly, and those with physical, psychosocial \\ and cognitive disabilities (i.e. mental illness and intellectual disability)\end{tabular} \\\hline
State Control & Including that of detainees, military conscripts, and persons in state institutions \\\hline
Victimisation & \begin{tabular}[c]{@{}l@{}}Due to victimisation, including by domestic and sexual abuse, other violations, \\ or because of a feeling of vulnerability\end{tabular} \\\hline
Migration & In the migration context, applies to detention and expulsion of asylum-seekers \\\hline
Discrimination & \begin{tabular}[c]{@{}l@{}}Due to discrimination and marginalisation, which covers ethnic, political and \\ religious minorities, LGBTQI people, and those living with HIV/AIDS\end{tabular} \\\hline
Reproductive Health & Due to pregnancy or situations of precarious reproductive health \\\hline
Unpopular Views & Due to the espousal of unpopular views \\\hline
Intersection & Intersecting vulnerabilities \\\hline
\end{tabular}
}

%% file: text/introduction.tex
Vulnerability encompasses a state of susceptibility to harm, or exploitation, particularly among individuals or groups who face a higher likelihood of experiencing adverse outcomes due to various factors such as age, health, disability, or marginalized social position \cite{mackenzieIntroductionWhatVulnerability2013a, finemanVulnerableSubjectAnchoring2016}.
% Recognizing and understanding vulnerability is essential as it enables us to identify those individuals or groups who may need additional support and protection to ensure their well-being. 
While it is impossible to eliminate vulnerability, society has the capacity to mitigate its impact. The European Court of Human Rights (ECtHR) interprets the European Convention of Human Rights (ECHR) to address the specific contextual needs of individuals and provide effective protection. This is achieved through various means, such as 
% narrowing the margin of appreciation, 
displaying flexibility in admissibility issues, and shifting the burden of proof \cite{heriResponsiveHumanRights2021}. 

\begin{figure}[h]
    \begin{subfigure}{.48\textwidth}
    \hspace{-5mm}
    \includegraphics[width=8.2cm]{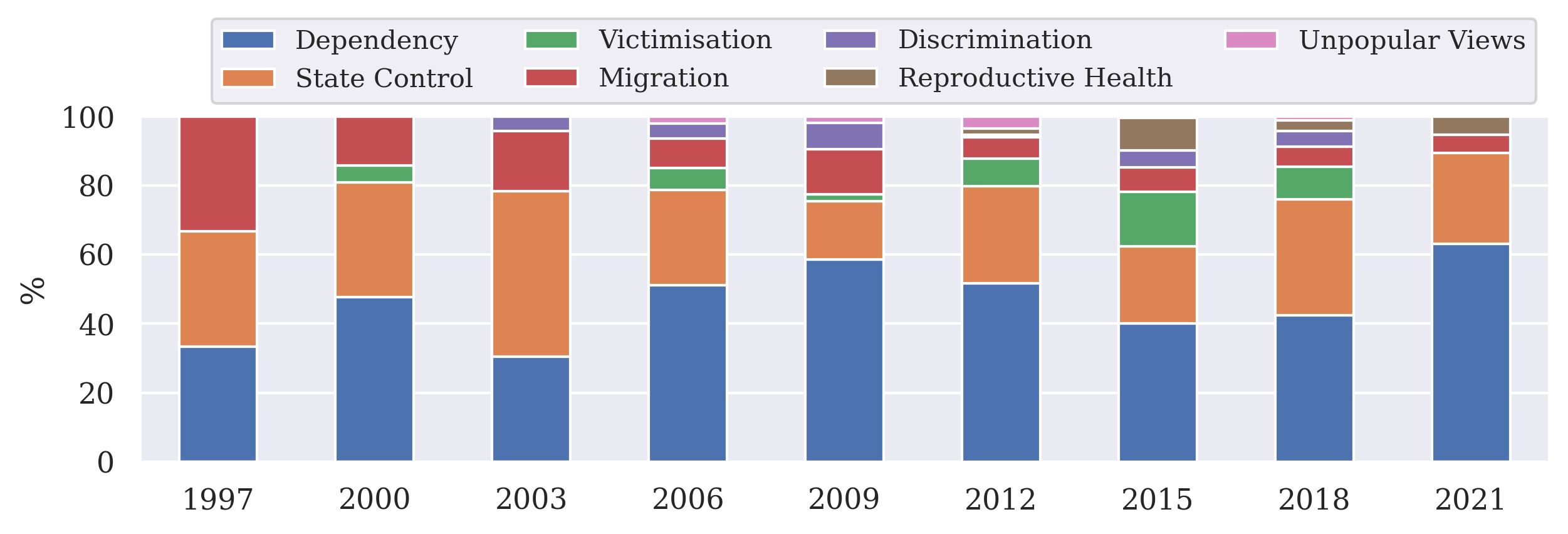}
    \caption{Evolving distribution of types of vulnerability.}
    \label{fig:distribution-temporal}
    \end{subfigure}
    
    \begin{subfigure}{.48\textwidth}
    \hspace{-5mm}
    \includegraphics[width=8.2cm]{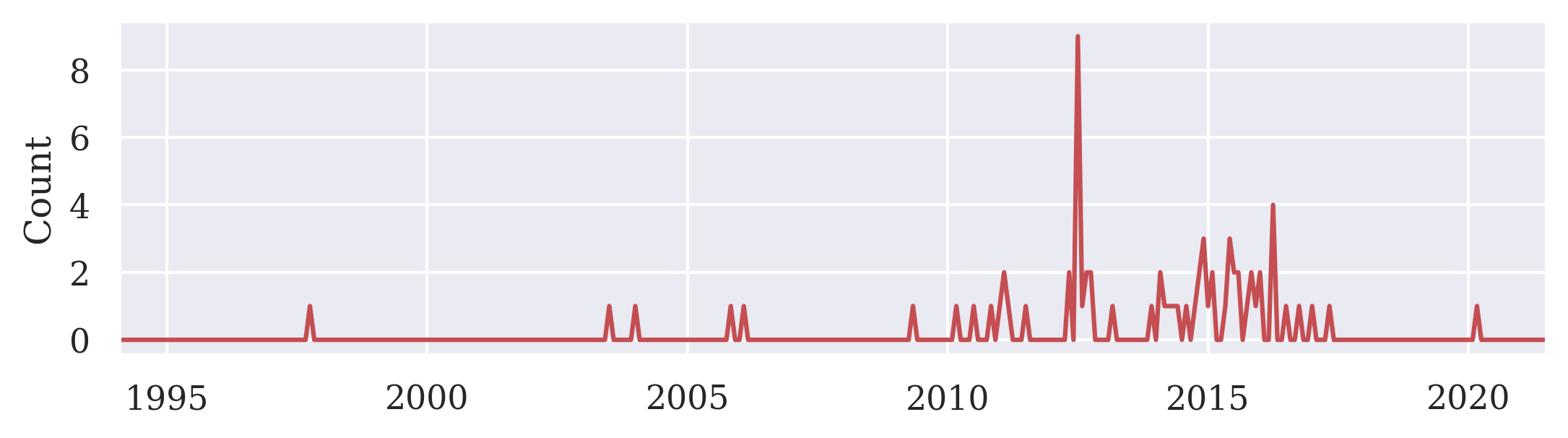}
    \caption{Increase in application of the vulnerability type \enquote{migration} between 2010 and 2018.}
    \label{fig:migration-temporal}
    \end{subfigure}
    \caption{Distribution changes of vulnerability types.} % TODO: perhaps remove again
\end{figure} 

However, the concept of vulnerability remains elusive within the ECtHR. While legal scholars have explored vulnerability as a component of legal reasoning \cite{peroniVulnerableGroupsPromise2013}, empirical work in this area remains scarce and predominantly relies on laborious manual processes. To address this challenge, NLP can offer valuable tools to assist experts in efficiently classifying and analyzing textual data. Besides high classification performance, the true utility of NLP in the legal field is its ability to identify relevant aspects related to vulnerability in court cases. These aspects can be extracted, grouped into patterns, and used to inform both litigation strategy and legal policy.
% In essence, the case outcome signal can be employed to construct the legal \enquote{concepts} of vulnerability. 
Even so, a significant obstacle to progress in this area is the lack of appropriate datasets. To bridge these research gaps, we present the dataset VECHR\footnote{VECHR stands for \textbf{V}ulnerability Classification in \textbf{E}uropean \textbf{C}ourt of \textbf{H}uman \textbf{R}ights. Our dataset and code is available at \url{https://github.com/TUMLegalTech/vechr_emnlp23}}, which comprises cases dealing with allegation of Article 3 \enquote{Prohibition of torture} and is obtained from legal expert’s empirical study\footnote{\citealt{heriResponsiveHumanRights2021}. Heri is the fifth author of this work.}. Our proposed task is to identify which type of vulnerability (if any) is involved in a given ECtHR case.

% Previous work \cite{chalkidis2021paragraph} investigated the explainability of ECtHR allegation prediction on an annotated dataset of 50 cases by identifying the allegation-relevant paragraphs of judgment facts sections and found fairly limited alignment between models and experts rationale. \citealt{santosh2022deconfounding} show that these models are drawn to shallow spurious predictors if not deconfounded. Both works derived and evaluated the explanation rationales at the fairly coarse paragraph level, which may over-estimate alignment. The RaVE dateset \cite{xu2023disagreement} contains explanations annotated by two legal experts at token-level, where limited agreement between experts’ annotation, as well as the misalignment of model and experts’ rationale are observed. 
As model explainability is crucial for establishing trust, we extend the dataset with VECHR\textsubscript{explain}, a token-level explanation dataset annotated by domain experts on a subset of VECHR. Its fine-grained token-level design mitigates performance overestimation of explainability when evaluated at the coarse paragraph level, as shown in previous works \cite{chalkidis2021paragraph, santosh2022deconfounding, xu2023disagreement}. %that derived and evaluated explanations for case outcome classification in the ECtHR at coarse paragraph-level. 
Further, the understanding and application of vulnerability in court proceedings change over time, reflecting societal shifts and expanding to encompass a wider range of types (\autoref{fig:distribution-temporal}). The volume of cases also fluctuates significantly in response to social and political events (\autoref{fig:migration-temporal}). To evaluate the model's robustness against distribution shifts, we further collect and annotate an additional out-of-domain (OOD) test set from cases involving non-Article 3 allegations, called VECHR\textsubscript{challenge}.

%\\
%\\ % TODO: improve if there is no better line break
% It’s important to check whether the model is robust to such distribution shifts. 
% The VECHR dataset contains 788 ECHR cases under Article 3 manually annotated by expert. 
We present comprehensive benchmark results using state-of-the-art (SOTA) models, revealing limited performance in vulnerability type classification in VECHR. We assess the models' alignment with expert explanations in VECHR\textsubscript{explain}, and observe limited agreement. Experiment results on VECHR\textsubscript{challenge} indicate that, although incorporating description of the vulnerability type helps to improve the models' robustness, the performance remains low overall due to the challenges posed by the distribution shift. 
Our experiments underscore the difficulty of vulnerability classification in ECtHR, and highlight a need for further investigation on improve model accuracy, explainability, and robustness.

%% file: text/vulnerability.tex
% Contemporary legal scholars assert a close connection between vulnerability and equality \cite{finemanVulnerableSubjectAnchoring2016}. Vulnerability encompasses the physical presence of individuals and their dependence on external entities, notably the state \cite{nealNotGodsAnimals2012}. 
The inescapable and universal nature of vulnerability, as posited by \citet{finemanVulnerableSubjectAnchoring2016}, underscores its significance in legal reasoning. For instance, the European Union has acknowledged the concept by establishing a definition for vulnerable individuals \cite{Directive2013332013}. However, it remains undefined within the context of ECtHR.
% despite the court's adaptation of Convention standards to address individual needs and ensure the protection of rights. 
To facilitate an examination of vulnerability and its application within the ECtHR, it is crucial to establish a typology recognized by the Court. Several scholars have endeavored to effectively categorize vulnerability for this purpose \cite{timmerQuietRevolutionVulnerability2016,limanteVulnerableGroupsCase2022}.  
% peroniVulnerableGroupsPromise2013
One notable study is conducted by \citet{heriResponsiveHumanRights2021}, which provides a systematic and comprehensive examination of the concept of vulnerability under ECHR Article 3. Heri proposes a complete typology encompassing eight types: \textit{dependency}, \textit{state control}, \textit{victimization}, \textit{migration}, \textit{discrimination}, \textit{reproductive health}, \textit{unpopular views} and \textit{intersections} thereof. \autoref{tab:description} gives a description for each type.

% Heri's typology serves as a vital legal theoretical foundation for our work and the development of the dataset. 

%% file: text/data_collection.tex
\subsection{Data Source and Collection Process}

\noindent\textbf{VECHR} consists of 788 cases under Article 3, which were collected based on \citeauthor{heriResponsiveHumanRights2021}'s study of the Court's case law references of vulnerability. 
% We are grateful to Heri for generously sharing her protocol with us. 
See \autoref{app:sampling} for details on Heri's case sampling methodology and our post-processing procedures. We divided the dataset chronologically into three subsets: training (--05/2015, 590 cases), validation (05/2015--09/2016, 90 cases) and test (09/2016--02/2019, 108 cases). 

\noindent\textbf{VECHR\textsubscript{explain}}: We selected 40 cases (20 each) from the val and test splits for the explanation dataset. Within each split, our sampling procedure involved two steps. First, we ensured coverage of all seven types by sampling one case for each type. Subsequently, we randomly selected an additional 13 cases to supplement the initial selection. 

\noindent\textbf{VECHR\textsubscript{challenge}}: To test the model’s ability to generalize across distribution shifts, we extend VECHR by collecting and annotating additional cases \emph{not} related to Article 3. Following Heri's method, we used the regular expression \enquote{vulne*} to retrieve all English relevant documents from the ECtHR’s public database HUDOC\footnote{\url{https://hudoc.echr.coe.int}} and exclude cases related to Article 3. We restricted the collection to the time span from 09/2016 (corresponding to start time of the test set) to 07/2022. In cases where multiple documents existed for a given case, we selected only the most recent document, resulting in a dataset consisting of 282 judgments. VECHR\textsubscript{challenge} can be regarded as an out-of-domain topical (OOD) scenario. The in-domain train/val/test of VECHR are all from the same text topic cluster of Article 3. The OOD VECHR\textsubscript{challenge} consists of non-Article 3 cases from different topic clusters (e.g. Article 10: freedom of expression), which involves different legal concepts and language usage.\footnote{For example, the Court recognizes the vulnerability of an elderly woman and provides her with protection under Article 3 (prohibiting torture) rather than Article 10 (freedom of expression)}
% \footnote{For instance, the court consider a elder female vulnerable and address to her special need for proctection under Art 3 (prohibiting torture) but not under under Article 10 (freedom of expression) }  
% e.g. Article 5 - Right to liberty and Security, Article 10 - Freedom of expression)
% This dataset presents unique challenges for NLP models due to two primary reasons. First, the inclusion of cases falling under different articles, which involves legal concepts and contexts different from that of Article 3. Second, the incorporation of documents spanning from February 2019 onwards (the end time of the test set), introducing a temporal shift and potentially novel legal issues.

\subsection{Vulnerability Type Annotation} We follow the typology and methodology presented by \citealt{heriResponsiveHumanRights2021}. She considered cases as \enquote{vulnerable-related}, only when \enquote{vulnerability had effectively been employed by the Court in its reasoning}. These cases are further coded according to the trait or situation (vulnerable type) giving rise to the vulnerability. In situations where the Court considered that multiple traits contributed to the vulnerability, she coded the case once for each relevant category. The resulting dataset comprises 7 labels\footnote{See \autoref{app:omit-intersectionality} for our justification for excluding the type \enquote{intersectionality}.}. Cases in which vulnerability was used only in its common definition, e.g. “financially vulnerability”, were regarded as ‘non-vulnerable’  and were labelled none 
 \begin{table}[htpb] %htpb
  \centering
  \input{table/dataset_stat.tex}
  \footnotesize\caption{Dataset statistics for each split, with number of cases ($C$), number of non-vulnerable cases ($C_{\lnot V}$), mean tokens ($T$) per case, mean paragraphs per case, mean labels ($L$) per case, and mean labels per case when only considering positive vulnerability cases ($C_V$).}
  \label{tab:stat_dataset}
\end{table}
of the 7 types. See \autoref{app:vuln-related} for more details of the definition of ``vulnerable-related''. 

% consider not having a line break here
For cases under Article 3, we adopted the labelling provided by Heri's protocol. For VECHR\textsubscript{challenge}, we ask two expert annotators\footnote{See \autoref{app:expert-background} for annotators' background and expertise. \label{fn:expert-background}} to label the case following Heri's methodology\footnote{For the reason of why Heri confined her study to Article 3 and why the typology also applies to cases under other articles, please refer to \autoref{app:art3-justification}.}. Each annotator has annotated 141 cases. 

\noindent\textbf{Inter-Annotator Agreement} To ensure consistency with Heri's methodology, we conducted a two-round pilot study before proceeding with the annotation of the challenge set (details in \autoref{app:pilot-study}). In each round, two annotators independently labelled 20 randomly selected cases under Article 3, and we compared their annotations with Heri's labels. The inter-annotator agreement was calculated using Fleiss Kappa, and we observed an increase from 0.39 in the first round to 0.64 in the second round, indicating substantial agreement across
%in a multi-label setting with 
seven labels and three annotators.

\subsection{Explanation Annotation Process}

The explanation annotation process was done using the GLOSS annotation tool \cite{savelka2018segmenting}, see \autoref{app:gloss} for details. Based on the case facts, the annotators was instructed to identify relevant text segments that indicate the involvement of a specific vulnerability type in the Court's reasoning. The annotators was permitted to highlight the same text span as an explanation for multiple vulnerable types.

%% file: table/dataset_stat.tex
%    Split &  \# Cases (C) &  Mean Tokens / C &  Mean Paragraphs /C &  Labels / C (with non-vulnerable cases) &  Labels / C (without vulnerable) \\
\resizebox{0.5\textwidth}{!}{
\begin{tabular}{lrrrrrr}
\hline
    Split &  $\#C$ & $T/C$ &  $P/C$ & $\#C_{\lnot V}$ &  $L/C$ &  $L/C_V$ \\
\hline
Train     &         590 &             5140 &                83 &                     325 &                                0.70 &                         1.55 \\
Validation       &          90 &             5077 &                77 &                      27 &                                1.41 &                         2.02 \\
Test      &         108 &             3992 &                57 &                      34 &                                1.13 &                         1.65 \\
Challenge &         282 &             4176 &                51 &                     144 &                                0.61 &                         1.24 \\
\hline
Total     &        1070 &             4765 &                72 &                     530 &                                0.78 &                         1.54 \\
\hline
\end{tabular}
}

%% file: text/dataset_analysis.tex
\begin{figure*}[htp]
\centering
% \begin{subfigure}[t]{.3\linewidth}
% \caption{Explanation-informed model for soft match (RQ2)}
%     \centering\includegraphics[width=\linewidth]{images/RQ2.png}
    
%   \end{subfigure}

  \begin{subfigure}[t]{.40\linewidth} 
    \centering\includegraphics[width=0.65\linewidth]{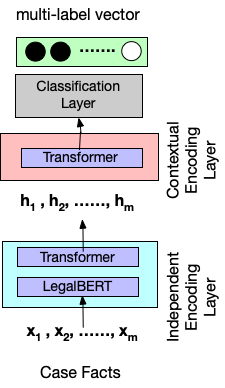}
     \caption{Hierarchical Classification Variant} \label{fig:arch_a}
    
  \end{subfigure}
  \hfill
  \begin{subfigure}[t]{.46\linewidth}

\centering\includegraphics[width=0.8\linewidth]{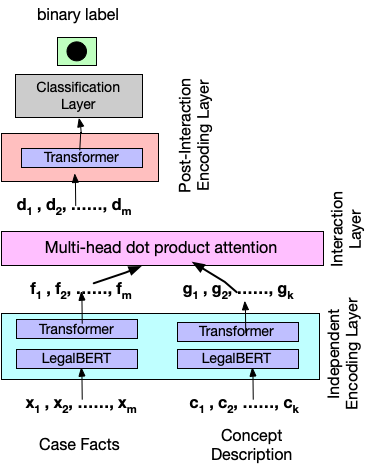}
  \footnotesize\caption{Concept-aware Classification Variant } \label{fig:arch_b}
    
  \end{subfigure}
  \caption{Visualization of Hierarchical and Concept-aware Hierarchical Model architectures.}
\label{fig:arch}
\end{figure*}

\autoref{tab:stat_dataset} presents the key statistics of our dataset. VECHR comprises a total of 1,070 documents, with an average of 4,765 tokens per case ($\sigma=4167$). 788 and 282 cases fall under the Article 3 and non-Article 3 partitions, respectively. Among all, 530 documents are considered as \enquote{non-vulnerable}, meaning they are not labelled as any of the seven vulnerable types. In the vulnerable-related cases, the average number of labels assigned per document is 1.54. 
% The distribution of label counts can be observed in \autoref{app:data_stat}. 

We observe a strong label distribution imbalance within the dataset. The label \enquote{state control} dominates, accounting for 33\% of the cases, while the least common label, \enquote{reproductive health}, is present in only 3\% of the cases. For more detailed statistics of our dataset, including details regarding the label imbalances in \autoref{tab:imbalance}, please refer to \autoref{app:data_stat}.

%% file: text/classification.tex
%\subsubsection{Models and Evaluation Metrics} 
\textbf{Task:} Our objective is to predict the set of specific vulnerability type(s) considered by the Court based on the factual text of a case. 

%Given the fact text of a case, we aim to predict which specific vulnerability type(s) are considered by the court. We provide results of various models for the proposed multi-label classification task. We consider the following models:\\

\noindent \textbf{Models:} 
We finetune pre-trained models
\emph{BERT} \cite{kenton2019bert}, \emph{CaselawBERT} \cite{zheng2021does}, \emph{LegalBERT} \cite{chalkidis2020legal}: on our dataset with a multi-label classification head, truncating the input to the maximum of 512 tokens.

% We use BERT (Devlin et al., 2019) (bert-uncased-base) as our baseline. 
% BERT has maximum input length of 512 tokens, whereas our case fact length  about 4.5K tokens. We truncate the documant to BERT’s maximum length, which affects its performance, as it might miss important information in later parts of the document. This also highlights an important limitation of BERT in processing long documents.

% \noindent\textbf{Law-specific BERT}: Legal-BERT (Chalkidis et al., 2020b) and CaseLaw-BERT (Zheng et al., 2021) are two law-specific BERTs. Both use the original pre-training BERT configuration. The LegalBERT pre-training corpora also consists of text from ECHR in addition to other corpus such as US, EU, UK legislation, contracts etc. CaseLawBERT is pre-trained exclusively from Harvard Law case corpusUS courts, consisting of only legal decisions from US courts. 

\begin{table}[tbp]
\input{table/results_cls_exp.tex}
\footnotesize\caption{Classification and explanation results. We report F1s for classification performance and Kappa score with standard error for explanation agreement.}
\label{tab:cls_exp}
\end{table}

% Beltagy et al., 2020 uses  sparse-attention to reduces the computational complexity of the model. It extends the maximum input limitation to 4090 tokens, enabling more comprehensive coverage of lengthy documents.
We finetune the \emph{Longformer} model \cite{beltagy2020longformer} on our dataset %pre-trained model 
that allows for processing up to 4,096 tokens, using a sparse-attention mechanism which scales linearly, instead of quadratically.

We further employ a \emph{hierarchical} variant of pre-trained LegalBERT to deal with the long input limitation. We use a greedy input packing strategy where we merge multiple paragraphs\footnote{Details and statistics on paragraphs are reported in \autoref{app:data_stat}.} into one packet until it reaches the maximum of 512 tokens. We independently encode each packet of the input text using the pretrained model and obtain representations ($h_{[CLS]}$) for each packet. Then we apply a non-pretrained transformer encoder to make the packet representations context-aware.
%, considering the surrounding packets.
Finally, we apply max-pooling on the  context-aware packet representations to obtain the final representation of the case facts, which is then passed through a classification layer. \autoref{fig:arch_a} illustrates the detailed architecture of the hierarchical model.
% See App \ref{app:arch} for detailed architecture on the model. 

For details on all models' configuration and training, please refer to App \ref{app:hyperparameter}.

\noindent\textbf{Evaluation Metrics:} 
% We employ macro F1 score as the main evaluation metric, driven by the imbalanced label distribution in our dataset. We believe each vulnerability type is equally important and should thus be weighed equally. Macro f1 makes sure the minority labels will not be overshadowed due to their limited representation. We also report micro f1 for completeness.
we report micro-F1 (mic-F1) and macro-F1 (macF1) scores for 7+1 labels, where 7 labels correspond to 7 vulnerability types under consideration and an additional augmented label during evaluation to indicate non-vulnerable.

\noindent \textbf{Results:} \autoref{tab:cls_exp} reports the results of classification performance. We observe that legal-specific pre-training improved the performance over general pre-training. However, BERT models still face the input limitation constraint. Both Longformer and Hierarchical models improved compared to truncated variants and are comparable to each other. Overall, we see low overall performance across models, highlighting the challenging task.

% Oberservation: 
% (1) Legal-specific BERTs perform better than plan BERT > legal knowledge helps (cite CaseHold)
% (2) Longformer and Hierachical performas even better: important to the wohle information of the text
% (3) Overall low performance: vul classification is a challegning task.

%% file: table/results_cls_exp.tex
\resizebox{0.48\textwidth}{!}{
\footnotesize
\begin{tabular}{|l|ll|l|}
\hline
Model & \multicolumn{2}{l|}{Classification} & Explanation \\ \hline
 & \multicolumn{1}{l|}{mac-F1} & mic-F1 & Kappa \\ \hline
random & \multicolumn{1}{l|}{19.02} & 25.07 & -0.11 ± 0.02 \\ \hline
BERT & \multicolumn{1}{l|}{24.31} & 41.78 & 0.02 ± 0.06 \\ \hline
CaselawBERT & \multicolumn{1}{l|}{27.31} & 45.16 & 0.04 ± 0.08 \\ \hline
LegalBERT & \multicolumn{1}{l|}{27.34} & 42.47 & 0.04 ± 0.07 \\ \hline
Longformer & \multicolumn{1}{l|}{31.49} & 46.21 & 0.11 ± 0.11 \\ \hline
Hierachical & \multicolumn{1}{l|}{31.46} & 45.32 & 0.10 ± 0.08 \\ \hline
\end{tabular}

\label{tab:cls_exp}}

%% file: text/explanation.tex
We use Integrated Gradient (IG) \cite{sundararajan2017axiomatic} to obtain token-level importance from the model with respect to each vulnerable type under consideration. We max pool over sub-words to convert token-level IG scores into word-level scores, followed by a threshold-based binarization. \autoref{tab:cls_exp} reports explainability performance expressed as the average of Cohen's $\kappa$ between the models' focus and the experts' annotations for the test instances. We observe that the low explainability scores among different models reflect their trend in classification scores and also echo the challenging nature of the task.

% \subsubsection{Models and Evaluation Metrics}
% - Use IG
% - same model in classification
% \subsubsection{Results and Discussion}

%% file: text/cross_article.tex
\begin{table}[]

\input{table/results_cross.tex}
\footnotesize\caption{Results on the challenge dataset.}
\label{tab:cross}
\end{table}

We assess the robustness of models to distributional shift using the VECHR\textsubscript{challenge} and present the performance in \autoref{tab:cross}. Notably, we observe a drop in macro-F1 score on VECHR\textsubscript{challenge} compared to the test set. We attribute this to the models relying on suboptimal information about vulnerability types, which is primarily derived from the factual content rather than a true understanding of the underlying concept. To address this limitation, we propose a \textbf{Concept-aware Hierarchical} model that considers both the case facts and the description of vulnerability type to determine if the facts align with the specified vulnerability type\footnote{We cast the multi-label task into a binary classification setup by pairing the text with each vulnerability type. These binary labels are transformed into a  multi-label vector for performance evaluation, to produce a fair comparison to multi-label models on the same metric.}, inspired by \citealt{tyss2023zero}. 
We employ a greedy packing strategy as described earlier and use a hierarchical model to obtain the context-aware packet representations for each packet in the facts and concept description separately. Subsequently, we apply scaled-dot-product cross attention between the packet vectors of the facts (as Query) and concepts (as Keys and Values), generating the concept-aware representation of the facts section packets. A transformer layer is used to capture the contextual information of the updated packet vectors. Then we obtain the concept-aware representation of the case facts via max pooling and pass it through a classification layer to obtain the binary label. \autoref{fig:arch_b} illustrates the detailed architecture of the concept-aware model. For more details, see \autoref{app:arch}.

The concept-aware model exhibits increased robustness to distributional shift and  %has maintained a performance with long-text  models (Longformer and Hierarchical) on the test set, but 
shows an improvement on the challenge set, owed to the incorporation of the vulnerability type descriptions. 
Overall, our results show promise for the feasibility of the task yet indicate room for improvement.
% +++ place holder Entailment or Eco-valid. +++
% For our entailment experiments, we augment the dataset with the texts of the 10 articles in the label set copied from the publicly available ECHR convention document2. We formulate the entailment variant for both tasks. Given both the case fact statements and a particular article information, the model should predict the binary outcome of whether an article has been alleged to be violated by the claimant (task B) or found to have been violated by the court (task A). The entailment variant enables us to experiment with LJP in zero-shot transfer (i.e to determine violation/allegation of case facts with respect to articles which are not seen during training time). This setup can be considered to be domain adaptation, where we consider a ‘domain’ to be the determination of outcomes based on case facts with regard to a particular convention article (i.e., 10 convention articles form 10 domains). The objective is then to train a model on a source domain (seen articles) with the goal of performing well at test-timeon the target domain (unseen articles).

%% file: table/results_cross.tex
% Please add the following required packages to your document preamble:
% \usepackage[normalem]{ulem}
% \useunder{\uline}{\ul}{}
\resizebox{0.48\textwidth}{!}{
\footnotesize
\begin{tabular}{|l|cc|}
\hline
 Model             & \multicolumn{2}{c|}{VECHR\textsubscript{challenge}}               \\ \hline
         & \multicolumn{1}{c|}{mac-f1}         & mic-f1 \\ \hline
random        & \multicolumn{1}{c|}{12.75}          & 14.61  \\ \hline
BERT          & \multicolumn{1}{c|}{20.51}          & 43.48  \\ \hline
CaselawBERT   & \multicolumn{1}{c|}{24.55}          & 57.51  \\ \hline
LegalBERT     & \multicolumn{1}{c|}{22.60}          & 50.77  \\ \hline
Longformer    & \multicolumn{1}{c|}{25.24}          & 55.71  \\ \hline
Hierarchical   & \multicolumn{1}{c|}{{\ul 26.43}}    & 58.46  \\ \hline
Concept-aware Hierarchical & \multicolumn{1}{c|}{\textbf{33.11}} & 49.62  \\ \hline
\end{tabular}

% }
\label{tab:cross}
}

%% file: text/limitations.tex
In our task, the length and complexity of the legal text require annotators with a deep understanding of ECtHR jurisprudence to identify vulnerability types. As a result, acquiring a large amount of annotation through crowdsourcing is not feasible, leading to limited-sized datasets. Additionally, the high workload restricts us to collecting only one annotation per case. There is a growing body of work in mainstream NLP that highlights the presence of irreconcilable Human Label Variation\cite{plank-2022-problem,basile-etal-2021-need} in subjective tasks, such as natural language inference \cite{pavlick-kwiatkowski-2019-inherent} and toxic language detection \cite{sap-etal-2022-annotators}. Future work should address this limitation and strive to incorporate multiple annotations to capture a more and potentially multi-faceted of the concept of vulnerability.

This limitation is particularly pronounced because of the self-referential wording of the ECtHR \cite{fikfakWhatFutureHuman2021}. As the court uses similar phrases in cases against the same respondent state or alleging the same violation, the model may learn that these are particularly relevant, even though this does not represent the legal reality. In this regard, it is questionable whether cases of the ECtHR can be considered \enquote{natural language}. Moreover, the wording of case documents is likely to be influenced by the decision or judgement of the Court. This is because the documents are composed by court staff after the verdict. Awareness of the case's conclusion could potentially impact the way its facts are presented, leading to the removal of irrelevant information or the highlighting of facts that were discovered during an investigation and are pertinent to the result \cite{medvedevaJudicialDecisionsEuropean}. Instead, one could base the analysis on the so-called ``communicated cases'', which are often published years before the case is judged. However, these come with their own limitations and only represent the facts as characterized by the applicant applicant and not the respondent state. There are also significantly fewer communicated cases than decisions and judgements.

% from santosh zero shot, rephrased
One of the main challenges when working with corpora in the legal domain is their extensive length. To overcome this issue, we employ hierarchical models, which have a limitation in that tokens across long distances cannot directly interact with each other. The exploration of this limitation in hierarchical models is still relatively unexplored, although there are some preliminary studies available (e.g., see \citealt{chalkidis2022exploration}). Additionally, we choose to freeze the weights in the LegalBERT sentence encoder. This is intended to conserve computational resources and reduce the model's vulnerability to superficial cues.

%% file: text/ethics.tex
Ethical considerations are of particular importance because the dataset deals with vulnerability and thus with people in need of special protection. In general, particular attention needs to be paid to ethics in the legal context to ensure the values of equal treatment, justification and explanation of outcomes and freedom from bias are upheld \cite{surdenEthicsArtificialIntelligence2019}. 

The assessment of the ethical implications of the dataset is based on the Data Statements by \citet{benderDataStatementsNatural2018}. Through this, we aim to establish transparency and a more profound understanding of limitations and biases. The curation is limited to the Article 3 documents in English. The speaker and annotator demographic are legally trained scholars, proficient in the English language. \enquote{Speaker} here refers to the authors of the case documents, which are staff of the Court, rather than applicants. We do not believe that the labelling of vulnerable applicants is harmful because it is done from a legally theoretical perspective, intending to support applicants. The underlying data is based exclusively on the publicly available datasets of ECtHR documents available on HUDOC\footnote{\url{https://hudoc.echr.coe.int}}. The documents are not anonymized and contain the real names of the individuals involved. We do not consider the dataset to be harmful, given that the judgments are already publicly available. 

We are conscious that, by adapting pre-trained encoders, our models inherit any biases they contain. The results we observed do not substantially relate to such encoded bias. Nonetheless, attention should be paid to how models on vulnerability are employed practically. 

In light of the aforementioned limitations and the high stakes in a human rights court, we have evaluated the potential for misuse of the vulnerability classification models. \citet{medvedevaDangerReverseEngineeringAutomated2020} mention the possibility of reverse engineering the model to better prepare applications or defences. This approach is, however, only applicable in a fully automated system using a model with high accuracy towards an anticipated decision outcome. As this is not the case for the models presented, we assume the risk of circumventing legal reasoning to be low. On the contrary, we believe employing a high recall vulnerability model could aid applicants and strengthen their legal reasoning. In a scholarly setting focused on vulnerability research, we do not think the model can be used in a detrimental way. Our research group is strongly committed to research on legal NLP models as a means to derive insight from legal data for purposes of increasing transparency, accountability, and explainability of data-driven systems in the legal domain.

There was no significant environmental impact, as we performed no pre-training on large datasets. Computational resources were used for fine-tuning and training the models, as well as assessing the dataset. Based on partial logging of computational hours and idle time, we estimate an upper bound for the carbon footprint of 30 kg $\mathrm{CO_2}$ equivalents. This is an insignificant environmental impact.

%% file: text/appendix.tex
% Article

\section{Vulnerability Typology and Descriptions}
\label{sec:appendix-descriptions}
\input{text/appendix_descriptions}

\section{More details on Data Source and Collection Process} \label{app:sampling}
\citealt{heriResponsiveHumanRights2021} reported the details of the case sampling process. The following serves as a summary of her case sampling methodology: She used the regular expression \enquote{vulne*} to retrieve all relevant English documents related to Article 3 from HUDOC, the public database of the ECtHR, excluding communicated cases and legal summaries, for the time span between the inception of the Court and 28 February 2019. This yielded 1,147 results.

Heri recorded her labeling in an Excel sheet, including the application number for each case. The application number serves as a unique identifier for individual applications submitted to the ECtHR. To create VECHR, we fetch all relevant case documents from HUDOC, including their metadata. During the post-processing, when one case has multiple documents, we keep the latest document and discard the rest, which yields 788 documents. 

\section{Definition of ``Vulnerable-related''} \label{app:vuln-related}
\citet{heriResponsiveHumanRights2021} specified that only cases where ``vulnerability had effectively been employed by the Court in its reasoning'' are regarded as ``vulnerable-related''. Cases in which vulnerability was used only in its common definition, or used only in the context of other ECHR rights, or by third parties, were excluded. 
To ensure clarity and alignment with Heri's perspective, we communicated with her during the annotation process to clarify the definition of ``vulnerable-related''. As a result, we determined that vulnerability is labeled (primarily) in situations where:

\begin{itemize}
    \vspace{-0.2cm}
    \item Vulnerability is part of the Court’s main legal reasoning.
    \vspace{-0.2cm}
    \item The alleged victims (or their children) exhibit vulnerability.
    \vspace{-0.2cm}
    \item The ECtHR, rather than domestic courts or other parties, consider the alleged victims vulnerable. 
\end{itemize}

\section{Omitting \enquote{Intersectionality} Label}
\label{app:omit-intersectionality}
The vulnerability type \enquote{intersectionality} was omitted because of its unclear usage in cases. Even more than the other typologies, it is a highly nuanced concept without a clear legal definition. Furthermore, \citet[117]{heriResponsiveHumanRights2021} says that the ECtHR does not engage with the concept of intersectionality in this form. Given that there are only 11 cases exclusively annotated as \enquote{intersectional} (out of a total of 59), the effect of disregarding it in this work is negligible. Omission does not suggest that intersectionality fails to play a role in the reasoning of the ECtHR or that we deem it irrelevant. Instead, we suggest exploring the concept of vulnerability as a combination of the other seven vulnerabilities.

\section{Annotator Background \& Expertise} \label{app:expert-background}
Two annotators performed the classification task. Annotator 1 (the fourth author) is a Post-doctoral Researcher at a European Research Centre, who has worked at the European Court of Human Rights as a case lawyer. Annotator 2 (the second author) is a law student, with also a philosophy and computer science background. Prior to annotation, both annotators had read Heri's book, and Annotator 2 had received a training session on the ECHR from Annotator 1. 

The explanation dataset was annotated by Annotator 1. 

% figures previous

\section{Justification of Article Applicability} \label{app:art3-justification}
Heri's \citeyearpar{heriResponsiveHumanRights2021} typology is limited to Article 3 of the ECHR, which pertains to the Prohibition of Torture. Under Article 3, the concept of vulnerability was first coined by the ECtHR, given that it deals with inhuman, degrading treatment and torture, which represent prototypical contexts of vulnerability. As such, an initial exploration under Article 3 is reasonable. 

Applying Heri's procedure to non-Article 3 cases is nonetheless justified according to our legal expert because Heri's underlying typology is based on \citet{timmerQuietRevolutionVulnerability2016} and relates to all articles. Furthermore, vulnerability is now a concept that is not limited to a single article, and which the ECtHR applies across articles. 

%Heri based on Timmer. Timmers was built to apply across all articles. court itself applies the concept of vulnerability across articles. vulnerability is now a sort of norm that no longer limited to a single article because the court applies in different contexts and articles. 

%  we have drafted succinct descriptions.

\section{Pilot study for Annotating Non-Article 3 Cases} \label{app:pilot-study}
In the first round, both annotators independently labeled 20 randomly selected cases under Article 3. After completing the labeling process, they compared their annotations with Heri's labels and engaged in a discussion to address any discrepancies and clarify their understanding of the vulnerability concept. In the second round, the annotators independently labeled another 20 randomly selected cases. We calculated the inter-annotator agreement using Fleiss Kappa to measure the consistency between Heri's labels and the annotations provided by our two annotators. The Fleiss Kappa agreement increased from 0.39 in the first round to 0.64 in the second round, which we consider to be substantial agreement in a multi-label setting involving seven vulnerable types and three annotators.

\section{GLOSS Annotation Tool} \label{app:gloss}
\begin{figure*}[p]
    \centering
    \includegraphics[width =0.98\textwidth]{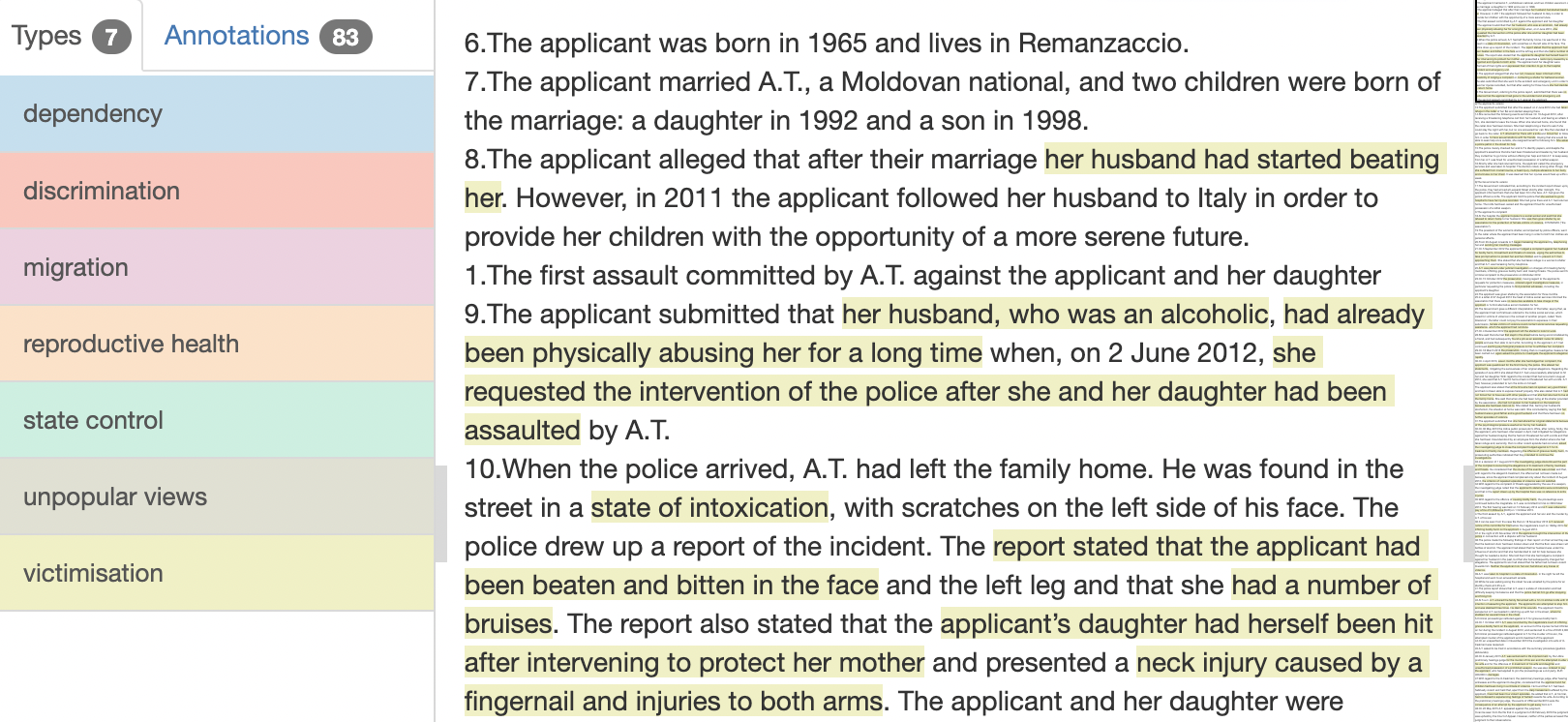}
    \caption{Screenshot of the GLOSS annotation interface.}
    \label{fig:gloss}  
\end{figure*}

The task of explanation annotation was done using the GLOSS annotation tool \cite{savelka2018segmenting}. \autoref{fig:gloss} demonstrates the GLOSS annotation interface. 

\begin{table}[htpb]%htpb
  \centering
  \input{table/explan_stat.tex}
  \caption{Statistics for the explanation dataset.}
  \label{tab:explan_dataset}
\end{table}

\begin{table}[htpb]
  \centering
  \input{table/imbalance_per_label}
  \caption{Count, percentage of total documents (\%), imbalance ratio per label (IRLbl) \cite{charteFirstApproachDeal2013} for each type of vulnerability.}
  \label{tab:imbalance}
\end{table}
%\vspace{-3mm} %% TODO: remove if there is a better solution

%%% Camera-Ready: move to main article
\section{Dataset Statistics} \label{app:data_stat}
\begin{figure}[htpb] %htpb
    \centering
    \includegraphics[width=8cm]{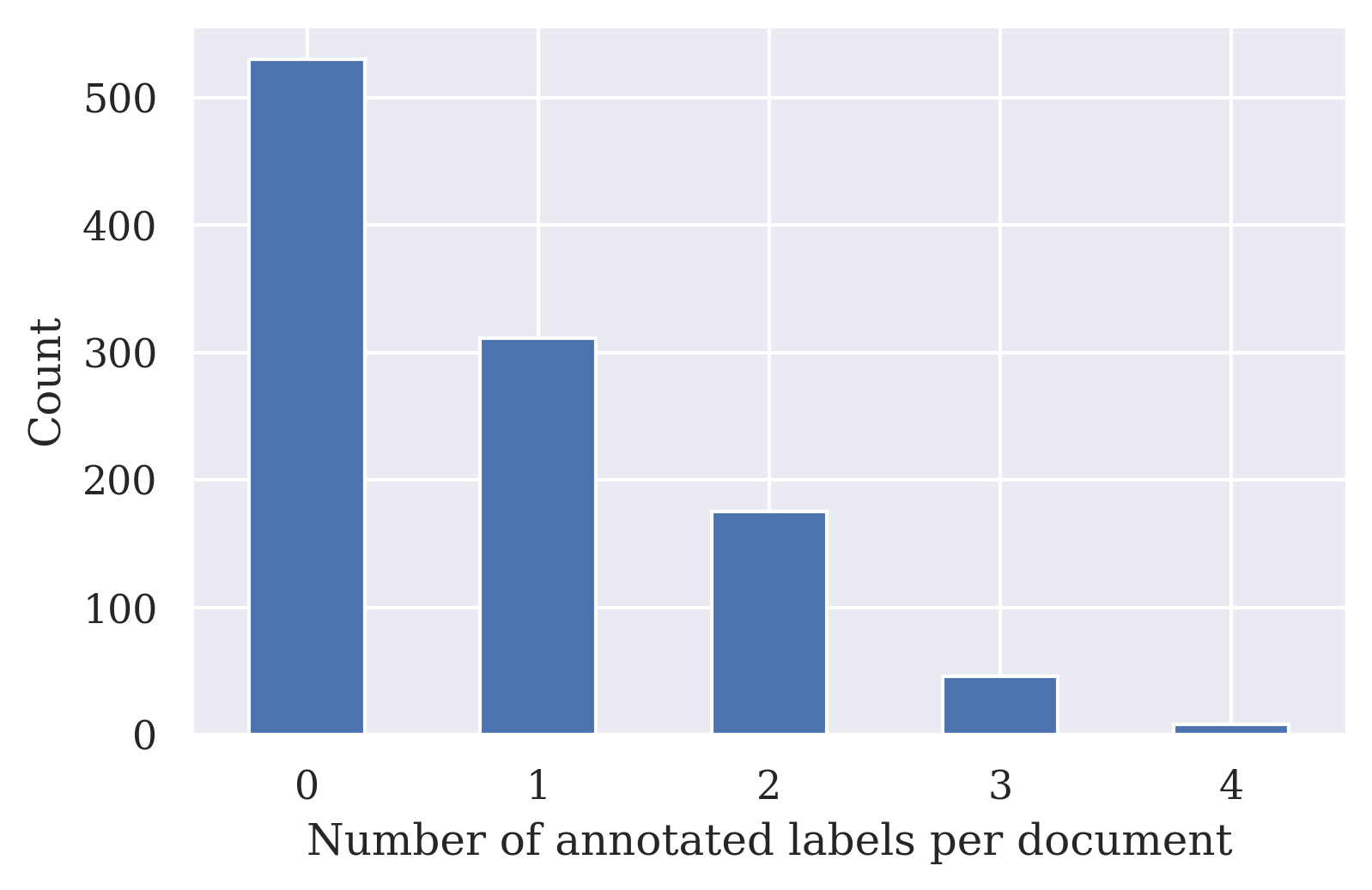}
    \caption{Distribution of number of annotated labels per document.}
    \label{fig:multi-count}
\end{figure}

\begin{figure}[htpbp]
    \centering
    \includegraphics[width=8cm]{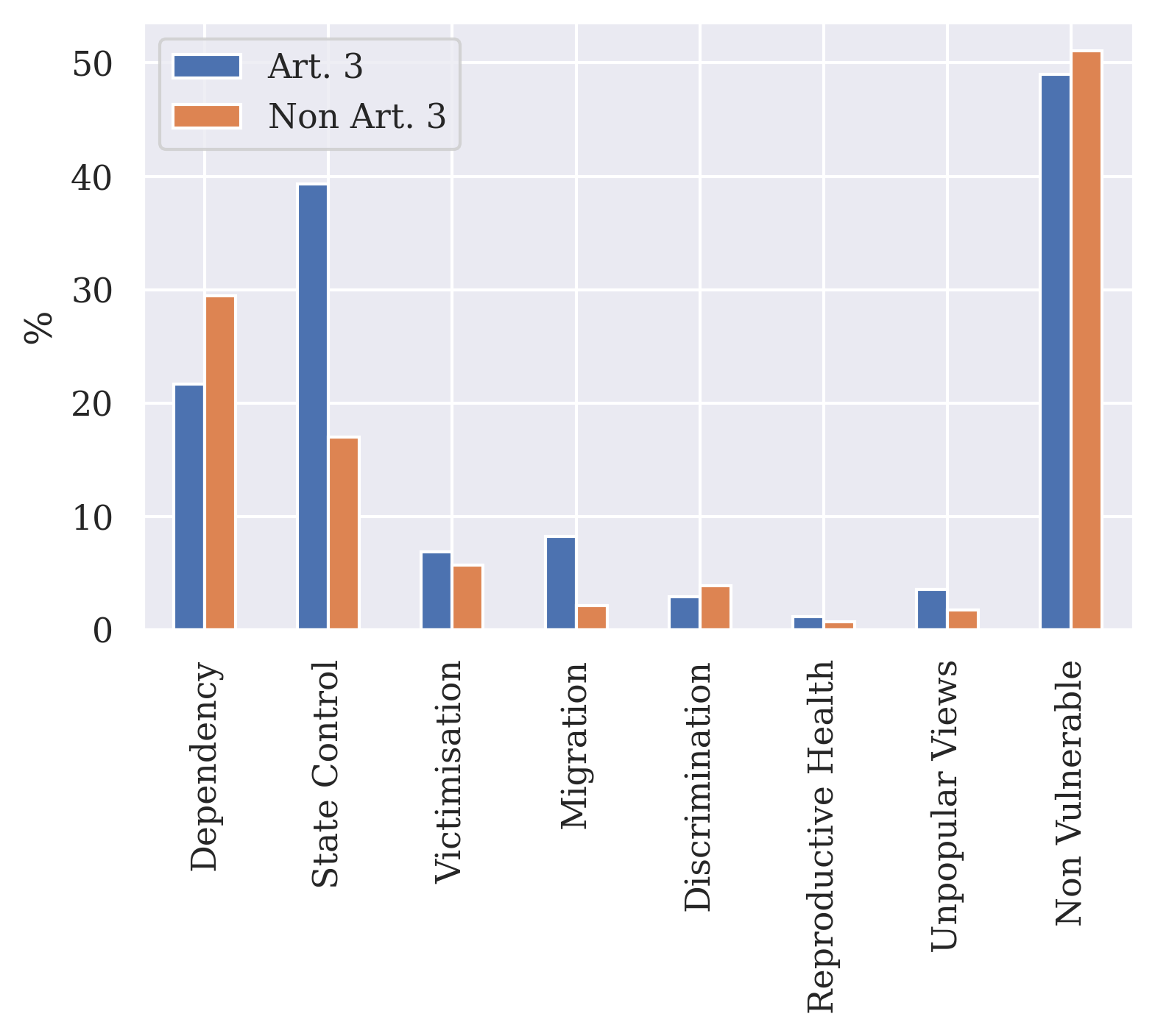} %width=8cm
    \caption{Difference in frequency of vulnerability annotations between Article 3 and non-Article 3.}
    \label{fig:annot-diff}
\end{figure}

\begin{figure}[htpbp] % htpb
    \centering
    \includegraphics[width=8cm]{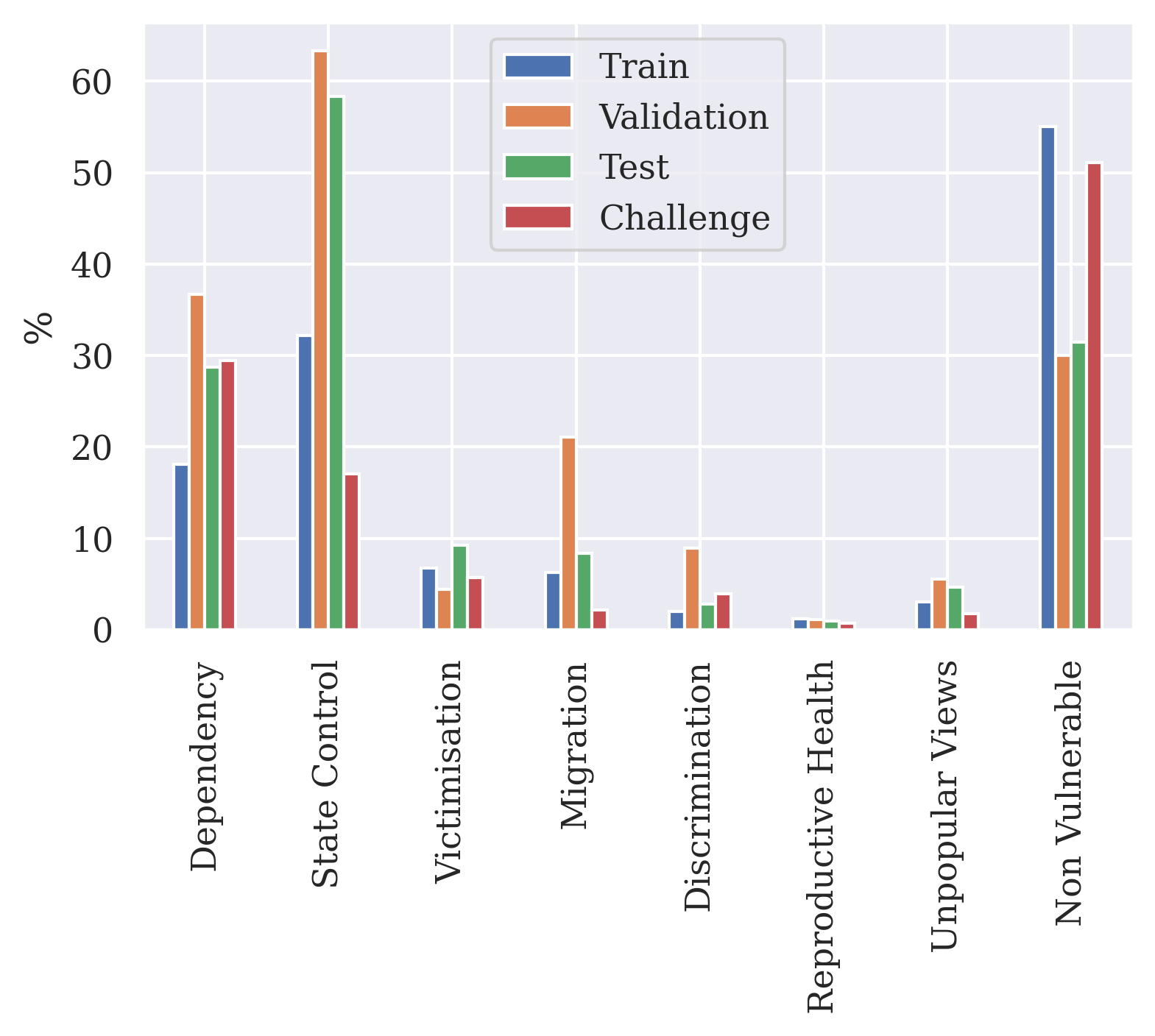}
    \caption{Difference in frequency of vulnerability type between \enquote{train}, \enquote{validation}, \enquote{test}, and \enquote{challenge} datasets.}
    \label{fig:annot-diff-all}
\end{figure}

The dataset comprises 1070 cases. On average, each case involves 0.78 vulnerable types and 1.54 vulnerable types for non-negative cases. 
\autoref{fig:multi-count} presents the distribution of the number of annotated labels per document. We report the imbalance characteristics for each label in \autoref{tab:imbalance}. \autoref{fig:annot-diff} shows the difference in frequency of vulnerability annotations between Article 3 and non-Article 3. \autoref{fig:annot-diff-all} illustrates the difference in frequency of each vulnerability label between all four datasets. \autoref{tab:explan_dataset} shows the statistics for the explanation dataset. 

\begin{figure*}
\centering
    
\begin{subfigure}[t]{\linewidth} %h!
    \centering
    \includegraphics[width=7.8cm]{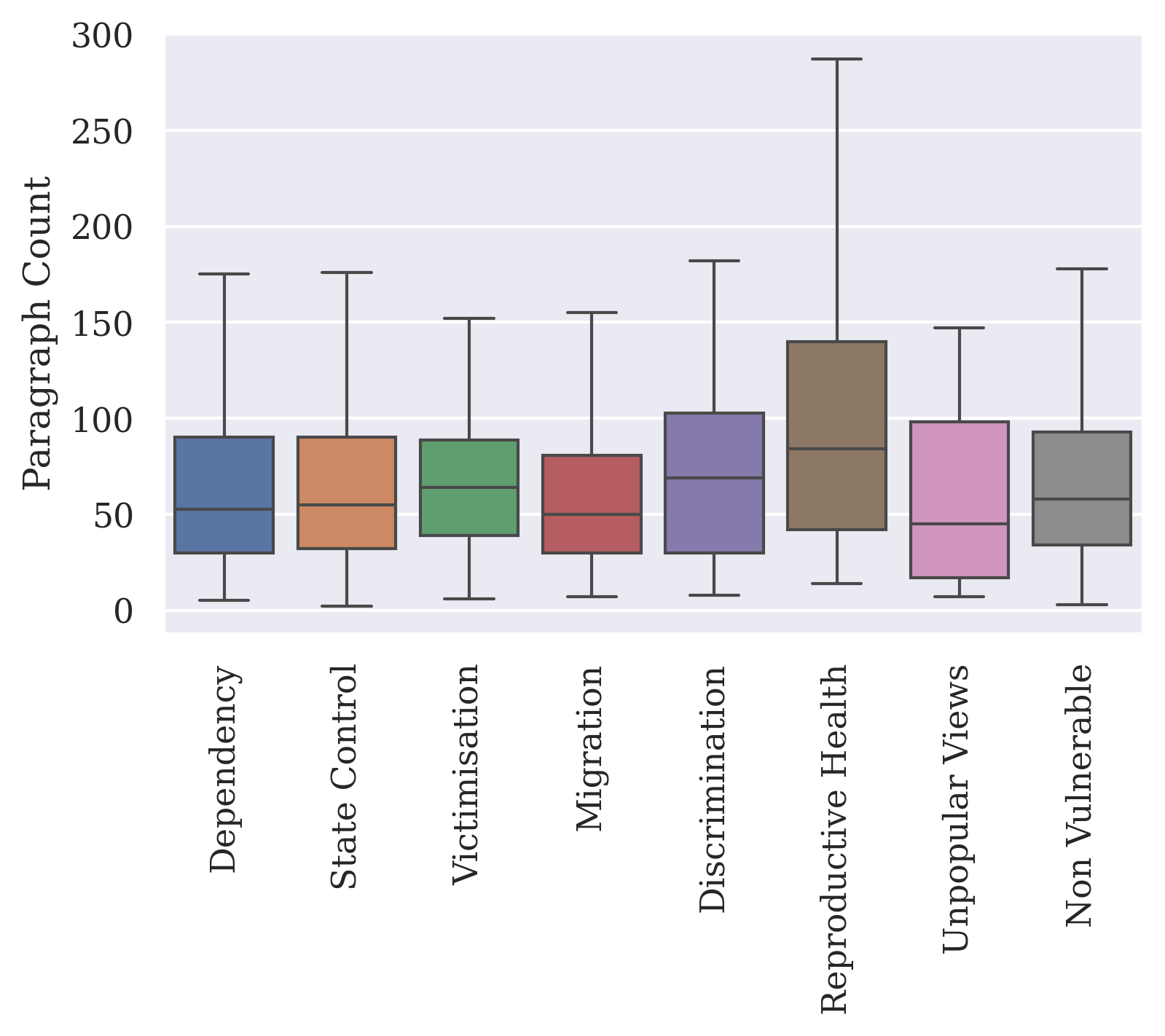}
    \subcaption{Box plot of the number of paragraphs per type.}
    \label{fig:dist-para}
\end{subfigure}

\begin{subfigure}[t]{\linewidth}
    \centering
    \includegraphics[width=7.8cm]{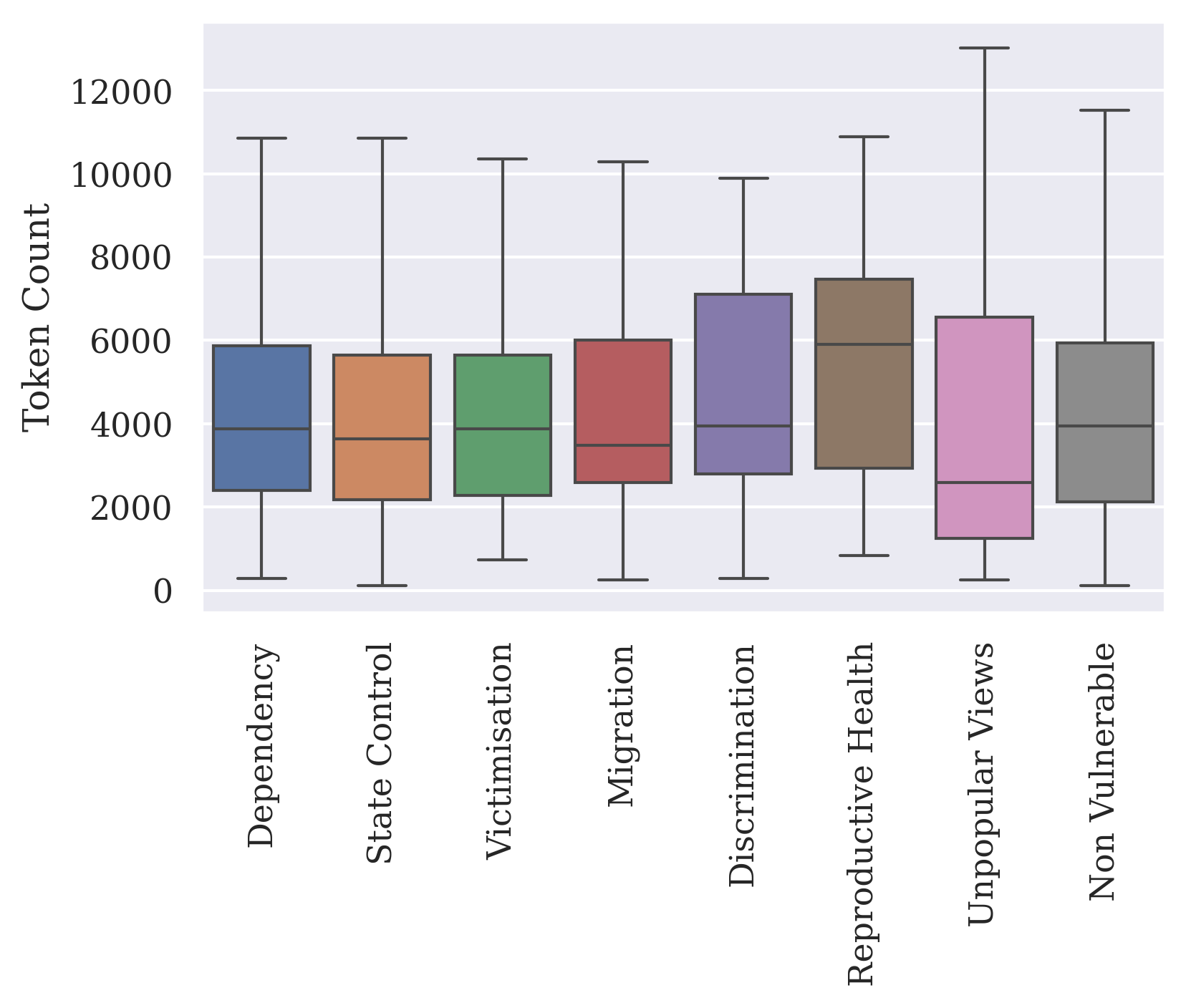}
    \subcaption{Box plot of the token count per type.}
    \label{fig:dist-token}
\end{subfigure}

\begin{subfigure}[t]{\linewidth} %h!
    \centering
    \includegraphics[width=7.8cm]{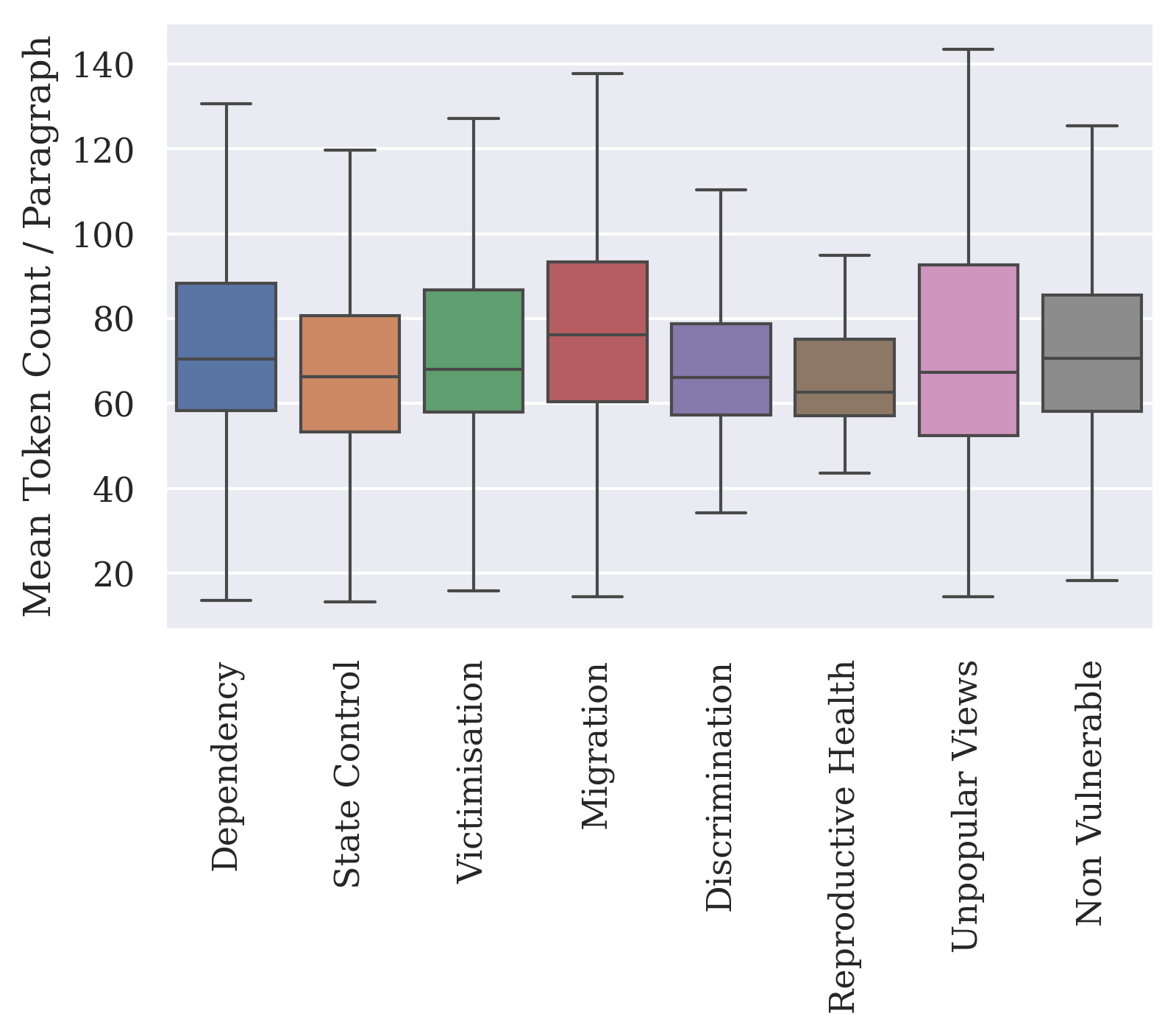}
    \subcaption{Box plot of the mean number of tokens per paragraph per type.}
    \label{fig:dist-token-para}
\end{subfigure}

\caption{Box plots on paragraph and token information for each vulnerability type.}

\end{figure*}

The hierarchical nature of the dataset is based on the naturally occurring paragraphs in the judgment texts. On average, each case consists of  71.54 paragraphs ($\sigma=67.54$). The distribution of the number of paragraphs by vulnerability type is shown in \autoref{fig:dist-para}. The mean token count is 4,765; its distribution by vulnerability type is depicted in \autoref{fig:dist-token}. The distribution of the mean token count per paragraph by vulnerability type is shown in \autoref{fig:dist-token-para}.
% Table \ref{tab:stat_dataset} presents the key statistics of our dataset. The dataset comprises 1070 cases, with an average of 4.5K tokens per case. On average, each case involves 1.5 vulnerable types. 
% \autoref{fig:multi-count} presents the distribution of the number of annotated labels per document. In \autoref{tab:imbalance} we report the imbalance characteristics for each label. \autoref{fig:annot-diff} demonstrates the difference in frequency of vulnerability annotations between Article 3 and non-Article 3. \autoref{tab:explan_dataset} shows the statistics for the explanation dataset.

% \begin{figure}[h]
%     \centering
%     \includegraphics[width=8cm]{emnlp2023-latex/figure/temporal-typols.png}
%     \caption{Difference in frequency of vulnerability annotations between article 3 and non- Article 3.
% }
%     \label{fig:temporal-typols}
% \end{figure}

\section{Implementation Details}\label{app:hyperparameter}
We use \emph{BERT} "bert-base-uncased" \cite{kenton2019bert}, \emph{CaselawBERT} "casehold/legalbert" \cite{zheng2021does}, \emph{LegalBERT} "nlpaueb/legal-bert-base-uncased" \cite{chalkidis2020legal}, and \emph{Longformer} "allenai/longformer-base-4096" \cite{beltagy2020longformer}. We finetune pre-trained models from the Transformers Hub \cite{wolf-etal-2020-transformers} on our dataset with a multi-label classification head, truncating to maximum lengths of 512 and 4096 tokens for BERT and Longformer models, respectively.\\
\textbf{Hyperparameter \& Overfitting Measures}: For the BERT-based models, we perform a grid search for hyperparameters across the search space of batch size [4, 8, 16] and learning\_rate [5e-6, 1e-5, 5e-5, 1e-4]. We train models with the Adam optimizer up to 8 epochs. We determine the best hyperparameters on the dev set and use early stopping based on the dev set macro-F1 score. 
For the Hierarchical models, we employ a maximum sentence length of 128 and document length (number of sentences) of 80. The dropout rate in all layers is 0.1. We perform a grid search for the hyperparameters across the search space of batch size [2, 4] and learning\_rate [1e-6, 5e-6, 1e-5]. We train models with the Adam optimizer up to 20 epochs. We determine the best hyperparameters on the dev set and use early stopping based on the dev set macro-F1 score. We use \emph{PyTorch} \cite{NEURIPS2019_9015} 2.0.1.

\section{Model Architecture}\label{app:arch}

\textbf{Hierarchical Model:}
Greedy packing turns the case facts input into $m$ packets as $x = \{x_1, x_2, \ldots, x_m\}$, where packet $x_i = \{x_{i1}, x_{i2}, \ldots, x_{in}\}$ consists of $n$ tokens. We pass each packet $x_i$ independently into the pre-trained LegalBERT model \cite{chalkidis2020legal} to extract the $h^{cls}_i$ representation for each packet. All packet representations $h= \{h_1, h_2, \ldots h_m\}$, along with learnable position embeddings, are passed through a transformer encoder to make them aware of the surrounding context. These context-aware packets representations are then max pooled to obtain the final representation of the case facts, which then pass through a classification layer. Given the multi-label nature of the task, we employ a binary cross-entropy loss over each vulnerability type label. \autoref{fig:arch_a}
illustrates the detailed architecture of our hierarchical model, which is inspired by \cite{santosh2022deconfounding,tyss2023leveraging}.

\noindent\textbf{Concept-aware Hierarchical model:} We cast the multi-label task into a binary classification setup where we pair case facts with each vulnerability type to predict whether this vulnerability type was raised by court in the case from which the facts section stems. Note that we transform these binary labels into a multi-label vector for evaluation, to make a fair comparison with multi-label models. This is based on the article-aware outcome prediction setting in \citealt{tyss2023zero}.

The concept-aware model also takes the case facts as input, which after greedy packing form $m$ packets $x = \{x_1, x_2, \ldots, x_m\}$ and vulnerability concept description text $c = \{c_1, c_2, \ldots, c_k\}$ with $k$ packets. Packet $x_i = \{x_{i1}, x_{i2}, \ldots, x_{in}\}$ has $n$ tokens and packet $c_i = \{c_{i1}, c_{i2}, \ldots, c_{ip}\}$ has $p$ tokens.

Similar to the hierarchical model, we use a pre-trained LegalBERT model \cite{chalkidis2020legal} to encode each packet in the case facts and concept description independently, and extract the $h^{cls}$ representation for each packet. These are passed through non-pretrained transformer model to obtain context-aware representations $f$ = $\{f_1,f_2,\ldots, f_m$\} and $g = \{g_1, g_2, \ldots g_k$\} for case facts and concept descriptions, respectively.

The obtained packet representations of facts and concept description interact with each other using a multi-head scaled dot product cross attention \cite{vaswani2017attention} similar to the decoder in the transformer layer by treating case facts packets ($f$) as the queries (Q) and the keys (K) and values (V) come from the concept description packets ($g$).
\begin{equation}
    \operatorname{Attention}(Q,K,V) = \operatorname{softmax}(\frac{QK^\top}{\sqrt{d_k}}\bigg)V
\end{equation}
Thus we obtain a concept-aware representations of the fact description packets $d = \{d_1,d_2,\ldots, d_m\}$, which are once again passed through non-pretrained transformer module to enhance them with the surrounding contextual information, and max-pooled operation to obtain the final concept-aware case fact representation. A classification layer then predicts the binary label indicating whether these facts give rise to the given vulnerability type. \autoref{fig:arch_b} displays the architecture of the concept-aware hierarchical model in detail.

%% file: text/appendix_descriptions.tex
Here is the typology of vulnerability in ECtHR \cite{heriResponsiveHumanRights2021}:
\begin{itemize}
    \item \textbf{Dependency}: dependency-based vulnerability, which concerns minors, the elderly, and those with physical, psychosocial and cognitive disabilities (i.e., mental illness and intellectual disability).
    \item \textbf{State Control}: vulnerability due to state control, including vulnerabilities 
    %that 
    of detainees, military conscripts, and persons in state institutions.
    \item \textbf{Victimisation}: vulnerability due to victimisation, including by domestic and sexual abuse, other violations, or because of a feeling of vulnerability.
    \item \textbf{Migration}: vulnerability in the migration context, applies to detention and expulsion of asylum-seekers.
    \item \textbf{Discrimination}: vulnerability due due to discrimination and marginalisation, which covers ethnic, political and religious minorities, LGBTQI people, and those living with HIV/AIDS.
    \item \textbf{Reproductive Health}: vulnerability due to pregnancy or situations of precarious reproductive health.
    \item \textbf{Unpopular Views}: vulnerability due to the espousal of unpopular views.
    \item \textbf{Intersection}: intersecting vulnerabilities.
\end{itemize}

Following is a detailed description of each type:\\
\noindent{\textbf{Dependency}} Dependency-based vulnerability derives from the inner characteristics of the applicant and thus concerns minors, elderly people, as well as physical, psychosocial and cognitive disabilities (i.e., mental illness and intellectual disability). The Court has built special requirements around these categories to be fulfilled by States as part of their obligations under the Convention.
Minors: The Court often refers to children as a paradigmatic example of vulnerable people and made use of the concept of vulnerability to require States to display particular diligence in cases imposing child protection given, on the one hand, their reduced ability and/or willingness of complaining of ill-treatment and, on the other hand, their susceptibility to be exposed to traumatic experiences/treatment.
Elderly: In many ways, vulnerability due to advanced age is a continuation of the vulnerability of children. All humans experience dependency at the beginning of life, and many experience it near the end.
Intellectual and Psychosocial Disabilities: Intellectual disability may render those affected dependent on others – be it the state or others who provide them with support. Having regard to their special needs in exercising legal capacity and going about their lives, the Court considered that such situations were likely to attract abuse. Persons living with intellectual disabilities experience difficulties in responding to, or even protesting against, violations of their rights. In addition, persons with severe cognitive disabilities may experience a legal power imbalance because they do not enjoy legal capacity.

\noindent{\textbf{State Control}} Vulnerability due to state control includes detainees, military conscripts, and persons in state institutions.
This type of vulnerability includes persons in detention, but also those who are institutionalised or otherwise under the sole authority of the state.
When people are deprived of their liberty, they are vulnerable because they depend on the authorities both to guarantee their safety and to provide them with access to essential resources like food, hygienic conditions, and health care.
In addition, the state often controls the flow of information and access to proof. Hence, the Court automatically applies the presumption of state responsibility when harm comes to those deprived of their liberty.

\noindent{\textbf{Victimisation}} Vulnerability due to victimisation refers to situations in which harm is inflicted by someone else. This type of vulnerability applies to situations of domestic and sexual abuse, and other type of abuse. The Court has also found that a Convention violation may, in and of itself, render someone vulnerable.  
Crime victims who are particularly vulnerable, through the circumstances of the crime, and can benefit from special measures best suited to their situation.
Sexual and domestic violence are expressions of power and control over the victim, and inflict particularly intense forms of trauma from a psychological standpoint.
Failing to recognise the suffering of the victims of sexual and domestic violence or engaging in a stigmatising response – such as, for example, the perpetuation of so-called ‘rape myths’ – represents a secondary victimisation or `revictimisation' of the victims by the legal system

\noindent{\textbf{Migration}} Vulnerability in the context of migration applies to detention and expulsion of asylum-seekers.
Applicants as asylum-seeker are considered particularly vulnerable based on the sole experience of migration and the trauma he or she was likely to have endured previously’.
The feeling of arbitrariness and the feeling of inferiority and anxiety often associated with migration, as well as the profound effect conditions of detention in special centres, indubitably affect a person’s dignity.
The status of the applicants as asylum-seekers is considered to require special protection because of their underprivileged (and vulnerable) status.

\noindent{\textbf{Discrimination}} Vulnerability due to discrimination and marginalisation covers ethnic, political and religious minorities, LGBTQI people, and those living with HIV/AIDS.
The Court recognises that the general situation of these groups – the usual conditions of their interaction with members of the majority or with the authorities – is particularly difficult and at odds with discriminatory attitudes.
Similarly to dependency-based vulnerability, this type of vulnerability imposes special duties on states and depends not solely on the inner characteristics of applicants but also on their choices which, in most cases, states have to balance against other choices and interests.

\noindent{\textbf{Reproductive Health}} Vulnerability due to pregnancy or situations of precarious reproductive health concerns situations in which women may find themselves in particular vulnerable situations, even if the Court does not consider women vulnerable as such.
This may be due to an experience of victimisation, for example in the form of gender-based violence, or due to various contexts that particularly affect women.
Sometimes, depending on the circumstances, pregnancy may be reason enough for vulnerability while other times vulnerability is linked to the needs of the unborn children.

\noindent{\textbf{Unpopular Views}} Vulnerability due to the espousal of unpopular views includes: demonstrators, dissidents, and journalists exposed to ill-treatment by state actors.
Where an extradition request shows that an applicant stands accused of religiously and politically motivated crimes, the Court considers this sufficient to demonstrate that the applicant is a member of a vulnerable group.
Similarly to the case of victimisation, it is also the applicant's choice to display such views and vulnerability comes from particular measures that state undertake when regulating or disregarding such choices.

%% file: table/explan_stat.tex
\resizebox{0.5\textwidth}{!}{
\begin{tabular}{lr}
\hline
\multicolumn{2}{c}{Case fact}                       \\ \hline
\# cases                         & 40               \\
Avg. \# vulnerable type per case & 1.3              \\
Avg. length per case             & 2964 ±1991 words \\ \hline
\multicolumn{2}{c}{Rationals from annotator 1}      \\ \hline
Avg. length case-allegation pair & 630 ± 551 words  \\ \hline

\end{tabular}
}
% \resizebox{0.5\textwidth}{!}{
% \begin{tabular}{lrrrrrr}
% \hline
%     Split &  \# C & T/C &  P/C & \# $\lnot$CV &  L/C &  L/CV \\
% \hline
% Train     &         590 &             5140 &                83 &                     275 &                                0.70 &                         1.49 \\
% Val       &          90 &             5077 &                77 &                      75 &                                1.41 &                         2.02 \\
% Test      &         108 &             3992 &                57 &                      63 &                                1.13 &                         1.63 \\
% Challenge &         282 &             4176 &                51 &                     138 &                                0.61 &                         1.24 \\
% \hline
% Total     &        1070 &             4596 &                67 &                     551 &                                0.96 &                         1.59 \\
% \hline
% \end{tabular}
% }

%% file: table/imbalance_per_label.tex
\resizebox{0.48\textwidth}{!}{
\begin{tabular}{lrrr}
\hline
 & Count & \% & IRLbl \\
\hline
Dependency & 254 & 23.74 & 1.41 \\
State Control & 358 & 33.46 & 1.00 \\
Victimisation & 70 & 6.54 & 5.11 \\
Migration & 71 & 6.64 & 5.04 \\
Discrimination & 34 & 3.18 & 10.53 \\
Reproductive Health & 11 & 1.03 & 32.55 \\
Unpopular Views & 33 & 3.08 & 10.85 \\
Non-Vulnerable & 530 & 49.53 & 0.68 \\
\hline
\end{tabular}
}